\documentclass[10pt,twocolumn,letterpaper]{article}
\usepackage{cvpr}      

\usepackage[dvipsnames]{xcolor}

\definecolor{cvprblue}{rgb}{0.21,0.49,0.74}
\usepackage[pagebackref,breaklinks,colorlinks,citecolor=cvprblue]{hyperref}
\usepackage{multirow}
\usepackage{multicol}
\usepackage{threeparttable}
\usepackage{tabularx}
\usepackage{booktabs}
\usepackage{makecell}
\usepackage{colortbl}
\usepackage{graphicx}
\usepackage{amsmath}
\usepackage{bbding}
\usepackage{listings}
\usepackage{arydshln}
\usepackage[ruled,linesnumbered]{algorithm2e}
\usepackage{color}
\newcommand{\blue}[1]{\textcolor{black}{#1}}
\def\paperID{1205} 
\newcommand{\figref}[1]{Fig.~\ref{#1}}
\newcommand{\tabref}[1]{Table~\ref{#1}}
\newcommand{\eqnref}[1]{Eq.~(\ref{#1})}
\newcommand{\secref}[1]{Sec.~\ref{#1}}

\def\confName{CVPR}
\def\confYear{2024}
\def\modelName{MMDG}
\def\adName{U-Adapter}
\def\gradName{ReGrad}
\title{Suppress and Rebalance: Towards Generalized Multi-Modal Face Anti-Spoofing}

\author{Xun Lin$^{1}$\!, Shuai Wang$^{1}$\!, Rizhao Cai$^{2}$\!, Yizhong Liu$^{1}$\!, Ying Fu$^{3}$\!, Zitong Yu$^{4}$\thanks{Corresponding author.}\,\,, Wenzhong Tang$^{1}$\!, Alex Kot$^{2}$
\\
$^1$Beihang University ~\quad
$^2$Nanyang Technological University ~\quad
$^3$BIT ~\quad
$^4$Great Bay University \\
{\tt\small \{linxun, wangshuai\}@buaa.edu.cn ~\quad yuzitong@gbu.edu.cn}
}

\begin{document}
\maketitle

\begin{abstract}
Face Anti-Spoofing (FAS) is crucial for securing face recognition systems against presentation attacks. 
With advancements in sensor manufacture and multi-modal learning techniques, many multi-modal FAS approaches have emerged. However, they face challenges in generalizing to unseen attacks and deployment conditions. 
These challenges arise from 
(1) modality unreliability, where some modality sensors like depth and infrared undergo significant domain shifts in varying environments, leading to the spread of unreliable information during cross-modal feature fusion, 
and (2) modality imbalance, where training overly relies on a dominant modality hinders the convergence of others, reducing effectiveness against attack types that are indistinguishable sorely using the dominant modality.
To address modality unreliability, we propose the \textbf{U}ncertainty-Guided Cross-\textbf{Adapter} (\textbf{\adName}) to recognize unreliably detected regions within each modality and suppress the impact of unreliable regions on other modalities. 
For modality imbalance, we propose a \textbf{Re}balanced Modality \textbf{Grad}ient Modulation (\textbf{\gradName}) strategy to rebalance the convergence speed of all modalities by adaptively adjusting their gradients.
Besides, we provide the first large-scale benchmark for evaluating multi-modal FAS performance under domain generalization scenarios. Extensive experiments demonstrate that our method outperforms state-of-the-art methods. Source code and protocols will be released on \url{https://github.com/OMGGGGG/mmdg}.

\end{abstract}

\section{Introduction}
The remarkable success of deep neural networks has also brought concerns regarding their security vulnerabilities \cite{crzfas2,yu2022towards,yy-backdoor,lx-imd}. 
Especially for face recognition (FR) systems, widely used in applications like surveillance and mobile payment \cite{survey}, are susceptible to various face presentation attacks, including printed photos, video replays, and 3D wearable masks \cite{crzfas1}. 
These vulnerabilities challenge the security of FR systems, limiting their broader application. As a result, the development of face anti-spoofing (FAS) methods has become a research hot-spot to enhance the security of FR systems.

\begin{figure}[t]
	\setlength{\abovecaptionskip}{3pt}
	\setlength{\belowcaptionskip}{-12.5pt}
	\centering 
	\includegraphics[width=0.45\textwidth]{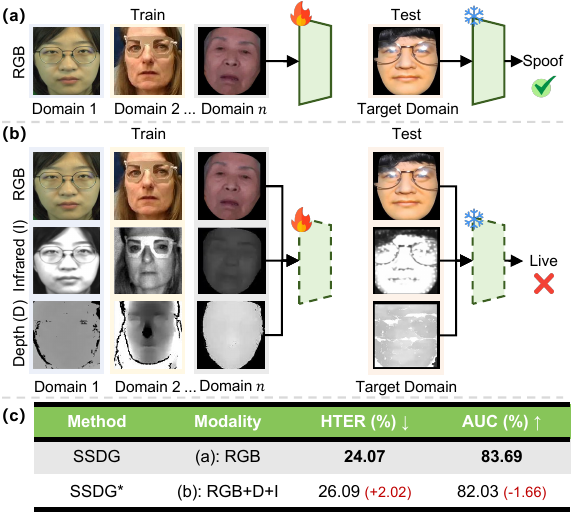}
 \vspace{-1mm}
	\caption{
 Illustration of DG scenarios in the context of (a) unimodal and (b) multi-modal. (c) DG performance of SSDG \cite{ssdg} on our Protocol 1 (see \secref{sec:bench}). Though fed with more modalities, SSDG performs worse in multi-modal scenarios compared to unimodal ones. * denotes our re-implemented multi-modal version. }
	\label{fig:first} 
 \vspace{-1.5mm}
\end{figure}

Recently, deep learning has been widely applied in FAS, particularly in unimodal scenarios based on the RGB spectrum. 
While these models perform well in intra-dataset evaluations, their generalizability to unseen attacks and deployment environments remains limited \cite{crzfas3}.
Despite the improvement of robustness via domain generalization (DG) techniques \cite{ssdg,iadg,ssan,lxm-cvpr23,crzfas4}, these approaches still tend to over-fit the attacks presented in source domains.

With the recent progress of multi-modal learning \cite{mm-survey,jyk-mm} and the development of sensor manufacture \cite{fm-benchmark}, some large-scale multi-modal FAS datasets \cite{cefa, wmca, mmfa, padisi} are proposed, which also drive the advancement in multi-modal FAS \cite{ma-vit, cmfl, mccnn, fm-vit, ama, mm-cdcn, vp-fas}. By expanding the input from RGB to multi-modal, \eg, RGB, depth, and infrared, richer spoofing traces can be captured for detecting various presentation attacks \cite{survey}. 
However, existing multi-modal methods fail to perform well under unseen deployment conditions. As illustrated in \figref{fig:first}, though the current methods generalize well in unimodal scenarios, their performance abnormally declines when extended and applied to multi-modal FAS.
\textit{Why using more modalities is not better?} We argue that there are two main reasons: 
(1) \textbf{Modality unreliability.} 
\blue{
Deploying in unseen environments or with various kinds of models of imaging sensors within the same modality, especially the depth and infrared shown in \figref{fig:first} (b), might introduce significant domain shifts \cite{mm-low-quality}.
Such shifts can result in unreliable multi-modal live/spoof feature extraction \cite{uc-physic}, misleading mutual modalities through cross-modal fusion.
}
(2) \textbf{Modality imbalance.} 
In some multi-modal learning scenarios, models tend to overly rely on a dominant modality with the highest convergence speed (\ie fast modality), which prevents itself from fully exploiting the other dominated modalities with relatively slower convergence speeds (\ie slow modalities) \cite{grad-cvpr22}.
In FAS, modality imbalance can cause significant negative impacts, especially when the fast modality is unsuitable or unavailable for detecting attacks in the target domain, relying on slow modalities that have not adequately converged cannot achieve satisfactory performance. 
This study addresses the cross-domain limitations of multi-modal FAS approaches due to modality unreliability and imbalance. 
As illustrated in Fig.~\ref{fig:overview}, we introduce a Multi-Modal Domain Generalized (MMDG) framework for FAS. Within MMDG, we develop trainable Uncertainty-guided cross-Adapters (U-Adapter) to fine-tune Vision Transformers (ViT) \cite{vit}
\blue{The \adName~address modality unreliability leveraging the uncertainty of each modality to suppress the focus on unreliable tokens during the cross-modal fusion. 
This prevents the excessive propagation of unreliable spoofing traces which tends to decrease the model's discriminative capability.}
Meanwhile, to solve modality imbalance, we design \textbf{Re}balanced modality \textbf{Grad}ient modulation strategy (\textbf{\gradName}). 
\blue{
Our \gradName~rebalance the convergence speed of all modalities by dynamically adjusting the backward propagated gradients of all trainable parameters in \adName s based on their conflict degree and convergence speed.}
This ensures that all modalities can be fully leveraged to resist various unseen attacks in the target domain. Our contributions are as follows.
\begin{itemize}
    \item \blue{
    We propose the MMDG framework to enhance the domain generalizability by addressing the modality unreliability and imbalance. This is the first investigation for multi-modal FAS under multi-source DG scenarios.
    }
	\item Within MMDG, we propose a \textbf{U}ncertainty-Guided Cross-\textbf{Adapter} (\textbf{\adName}) for ViT-based multi-modal FAS, which can complementarily fuse cross-modal features and prevents each modality from being influenced by unreliable information caused by domain shifts.
	\item We design the \textbf{Re}balanced Modality \textbf{Grad}ient Modulation (\textbf{\gradName}) for MMDG, a strategy that can monitor and rebalance the convergence speed of each modality by adaptively modulating gradients of \adName s.
	\item \blue{We build the first large-scale benchmark to evaluate the domain generalizability of multi-modal FAS approaches. Extensive experiments on this benchmark demonstrate the superiority and generalizability of \modelName.} 
\end{itemize}
\section{Related Works}
\subsection{Domain Generalization for FAS}
\blue{In FAS, domain generalization focuses on training a model using multiple source domains, with the intention that this model can work effectively on unseen target domains \cite{adaptive-transformer, dg2019}.}
Previous works verify the effectiveness of adversarial training \cite{trace-distanglement, conditional-feature, cyclically}, asymmetric triplet loss \cite{ssdg, divt} and controversial learning \cite{lxm-cvpr23, udg} in developing a shared feature space across domains.
Besides, some studies \cite{ssan, iadg, uda} investigate the style information of different domains, aiming to learn domain-invariant features by minimizing distinct style characteristics. 
Furthermore, meta-learning-based methods \cite{meta-teacher, dual-meta, meta-pattern, expert, ebdg} are also discussed to simulate domain shifts during training, fostering the learning of a robust representative feature space to such variations.
More recent works focus on parameter-efficient transfer learning \cite{continual, adaptive-transformer, flip, s-adapter}, enabling a pretrained ViT to efficiently adapt to unseen domains with a lower risk of over-fitting. 
However, these approaches are proposed for unimodal (\ie RGB) FAS. Due to the ignorance of modality unreliability, there is a gap in adapting them to multi-modal scenarios. Simple multi-modal extensions of these approaches, through early fusion, late fusion, or even the substitution with multi-modal backbones, still exhibit limited performance under multi-modal DG scenarios. 

\subsection{Multi-Modal FAS}
Multi-modal FAS refers to the use of multiple spectra, \eg, RGB, depth, and infrared, to reveal live and spoofing traces. Due to the complementary information among different modalities, traces that are undetectable in one modality can be easily captured in others \cite{fm-vit,hyperbolic}.
To fully exploit information from each modality, early methods employ channel-wise concatenation of inputted modalities \cite{mccnn,mc-face-pad}, or develop multiple separate branches for modality feature extraction followed by late fusion \cite{mm-cdcn,facebagnet, kcq-fas1, kcq-fas2}. 
Recent works introduce attention-based feature fusion techniques \cite{hd-aaai23,dual-stream} and adaptive cross-modal loss functions \cite{cmfl} to encourage the extraction of complementary information among modalities. 
Meanwhile, cross-modality translation is discussed \cite{cross-translate,lizhi} to mitigate the semantic gap between different modalities.
To further enhance the practical usability of FAS, \citet{fm-benchmark,vp-fas,ama} introduce flexible-modal benchmarks that force the use of incomplete modalities during training or testing. 
In the context of flexible-modal FAS, cross-modality attention \cite{fm-vit, ma-vit} or multi-modal adapters \cite{ama, vp-fas} are proposed for pretrained ViT to learn modality-agnostic live/spoofing features.
However, the aforementioned methods fail to perform well in DG scenarios. The lack of sensitivity to modality unreliability and imbalance leads to an insufficient ability to resist domain shifts. Although previous works rebalance modalities by modulating gradients \cite{grad-cvpr22} or adjusting modality losses \cite{pmr}, these methods ignore domain shifts and are not suitable for FAS frameworks with complex cross-modal fusion modules.

\begin{figure}[t]
	\setlength{\abovecaptionskip}{2.5pt}
	\setlength{\belowcaptionskip}{-12.5pt}
	\centering 
	\includegraphics[width=0.42\textwidth]{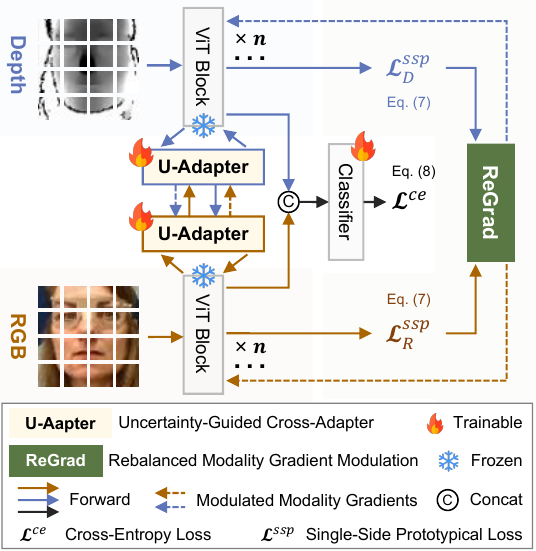}
	\caption{Overall structure of our \modelName. It consists of ViT-based backbones fine-tuned by the proposed \adName s with the modality rebalancing strategy called ReGrad. 
    Each modality is designed with a branch for feature extraction and enables feature interaction and mutual complementarity with other modalities. 
    For simplicity, we illustrate the two-modality scenario.  
    } 
	\label{fig:overview} 
 \vspace{-1mm}
\end{figure}

\section{Proposed \modelName}
As shown in \figref{fig:overview}, our \modelName~consists of ViT-based backbones fine-tuned by the proposed U-Adapters and ReGrad strategy, which are respectively designed for solving modality unreliability and imbalance.

\begin{figure*}[t]
	\setlength{\abovecaptionskip}{2.5pt}
	\setlength{\belowcaptionskip}{-11pt}
	\centering 
	\includegraphics[width=1.0\textwidth]{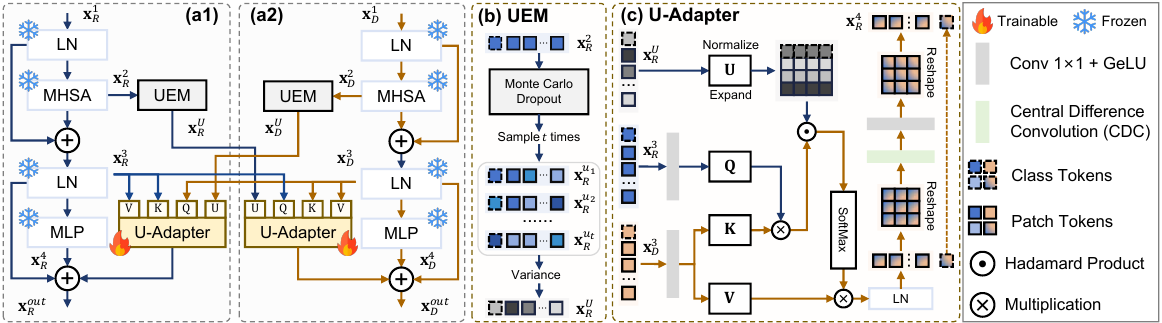}
	\caption{(a1)-(a2) Illustration of fine-tuning ViT with proposed U-Adapters, \blue{showcasing the interaction between the RGB (R) and Depth (D) modalities. Note that only parameters of U-Adapters are trainable.} (b) Uncertainty Estimation Module (UEM) used for recognizing unreliable tokens. (c) Detailed structure of \adName, which adopts cross-modal fusion and suppresses the interference of unreliable tokens on other modalities. After fusion, discriminative central difference information is integrated for fine-grained spoof representation.} 
	\label{fig:adapter} 
 \vspace{-1.5mm}
\end{figure*}

\subsection{Fine-tuning ViT with \adName s}
\blue{In multi-modal FAS, DG scenarios involving significant domain shifts \cite{uc-da}, including unseen face presentations attacks \cite{uc-physic}, different kinds of imaging sensors, and noise from low-quality sensors \cite{mm-low-quality}, often lead to the extraction of unreliable live/spoofing features from certain modal inputs. 
These unreliable features can propagate during cross-modal fusion, adversely affecting other modalities. 
To this end, we aim to identify unreliable local features within each modality and suppress the interaction of these features with other modalities, ensuring that each modality utilizes reliable information from other modalities as complementary traces.
Specifically, we adopt uncertainty estimated via Monte Carlo sampling \cite{mcdropout, bayesian, mcdropout-nn} to evaluate the feature unreliability in each modality, and based on the uncertainty, we reduce their weight in cross-modal attention-based feature fusion, minimizing their negative impact.}
Inspired by recent success in parameter-efficient transfer learning (PETL) by introducing adapters into pretrained ViT \cite{adaptive-transformer, continual, s-adapter, ama, vp-fas, luoadapter}, but distinctively, we propose the \adName~for fine-tuning, which enables cross-modal feature fusion and recognizes as well as suppresses the propagation of unreliable information from uncertainly detected regions caused by domain shifts.
In \figref{fig:adapter}, we present the process of fine-tuning ViT with the \adName s, including adopting an Uncertainty Estimation Module (UEM) to recognize unreliable tokens and employing the \adName~for uncertainty-aware cross-modal fusion.
The ViT \cite{vit}, comprising $n$ transformer blocks each with Layer Normalization (LN), Multi-Head Self Attention (MHSA), and Multi-Layer Perceptron (MLP), is fine-tuned at the MLP’s output using our U-Adapter. 
By representing our U-Adapter as $\mathcal{A}$ and taking RGB (R), depth (D), and infrared (I) as the inputted modalities, the fine-tuning process is formulated as follows.
\begin{equation}
    \footnotesize
    \mathbf x_R^{out} = \mathcal{A}(\mathbf x_D^U, \mathbf x_D^3, \mathbf x_R^3)\!+\!\mathcal{A}(\mathbf x_I^U, \mathbf x_I^3, \mathbf x_R^3) + \mathbf x_R^3 + \mathbf x_R^4,\\
\end{equation}
\begin{equation}
    \footnotesize
    \mathbf x_D^{out} = \mathcal{A}(\mathbf x_R^U, \mathbf x_R^3, \mathbf x_D^3) + \mathbf x_D^3 + \mathbf x_D^4,
\end{equation}
\begin{equation}
    \footnotesize
    \mathbf x_I^{out} = \mathcal{A}(\mathbf x_R^U, \mathbf x_R^3, \mathbf x_I^3) + \mathbf x_I^3 + \mathbf x_I^4,
\end{equation}
where for $m\in\{R,D,I\}$, $\mathbf x_m^U \in \mathbb R^{B\times L \times 1}$ are the estimated token uncertainty in the corresponding modality, and 
$\mathbf x_m^3, \mathbf x_m^4 \in \mathbb R^{B\times L \times C}$ are the outputs of the last LN and MLP, respectively.
The meaning of other vectors is illustrated in Fig. \ref{fig:adapter}. 
To prevent the modalities that experienced significant domain shifts, \ie depth and infrared, from misleading each other, we prohibit their direct interaction.

\vspace*{4pt}
\noindent\textbf{Uncertainty Estimation Module.}
As discussed above, it is essential to recognize unreliable modal information for FAS. 
\blue{In deep learning, uncertainty reflects the level of confidence a model has in its predictions. When the model is less certain about a prediction, we consider that prediction to be unreliable \cite{mcdropout}.}
When a model takes images with substantial noise caused by significant domain shifts as inputs, uncertainty estimation techniques \cite{mcdropout} tend to assign elevated uncertainty values to regions more likely to be misclassified \cite{uc-da}. 
One well-known approach is the Bayesian Neural Network (BNN), widely applied in various vision tasks \cite{uc-action,uc-imd}. BNN focuses on learning the posterior distribution of the model's weights rather than output a single value. 
Since the posterior distribution is always not tractable \cite{mcdropout}, approximate inference methods, such as Monte Carlo dropout (MCD) \cite{uc-med, uc-robot, mcdropout-nn}, are employed for estimation. 
In this work, we also use MCD to construct the UEM as a probabilistic representational model, which imposes a Bernoulli distribution over the weights of a model without introducing extra parameters.
As presented in \figref{fig:adapter} (b), we perform MCD instead of the vanilla dropout after each MHSA, simulating the effect of sampling from the posterior distribution by randomly dropping out neurons in MHSA.
According to Bayesian probability theory, we consider $\mathcal{T} =\{\mathbf x^{u_1}_R, \mathbf x^{u_2}_R,\cdots, \mathbf x^{u_{t}}_R \}$ as empirical samples from the approximate distribution, and use their token-wise variance $\mathbf x^{U}_R = \textrm {Var}(\mathcal{T}) \in \mathbb R^{B\times L\times 1}$ to reflect their unreliability.

\vspace*{4pt}
\noindent\textbf{Uncertainty-Guided Cross-Adapter.}
Different from existing cross-modal fusion \cite{ma-vit, fm-vit}, our \adName, guided by the estimated uncertainty, suppresses the interaction with unreliable tokens from other modalities to ensure the reliability of the fused features. 
As shown in \figref{fig:adapter} (c), \adName~interactively fuses features from different modalities and suppresses unreliable information during fusion. For example, in the depth branch (see \figref{fig:adapter} (a2)), \adName~takes the output of the second LN from another modality $\mathbf x_{R}^3$ as the \textit{query} for attention-based feature fusion. The output of the second LN from its own modality $\mathbf x_{D}^3$ is used as \textit{key} and \textit{value}. Moreover, \adName~modulates the \textit{query} tokens from the other modality based on their uncertainty. 
Since malicious presentation attacks result in subtle traces, we adopt central difference convolution (CDC) \cite{cdcn} to integrate the local details for ViT and explore fine-grained difference information from neighbor tokens, thus making the fused features more discriminative. By ignoring GeLU and vanilla convolution layers, the \adName~is formulated as follows.
\begin{equation}
	\footnotesize
        \mathcal{A}(\mathbf x_R^U,\!\mathbf x_R^3,\!\mathbf x_D^3)\!=\!
        \mathcal{C}\!\left(\!\mathcal{S}(\frac{\textrm Q(\mathbf x_R^3)\textrm K(\mathbf x_D^3)^\intercal\!\odot\!\textrm U(\mathbf x_R^U)}{\sqrt{n_k}})\textrm V(\mathbf x_D^3))\!\right),
	\label{eq:ad}
\end{equation}
where $\mathcal{C}(\cdot)$ denotes the CDC layer, $\mathcal{S}(\cdot)$ represents the softmax, and $n_k$ is the number of token channels. $\textrm Q(\cdot)$, $\textrm K(\cdot)$, and $\textrm V(\cdot)$ are the linear projections corresponding to the \textit{query}, \textit{key}, and \textit{value}, respectively. $\odot$ is the Hadamard product. 
\blue{$\textrm U(\mathbf x_R^U) = \textrm{exp}(-r_e\cdot \mathbf x_R^U \times \mathbf I)$ normalizes the inputted uncertainty using an exponential function, where $r_e$ is a penalty intensity parameter, and $\mathbf I=\left[1,1,\dots,1\right]_{1\times L}$ is adopted to expend the dimensions of $x_R^U$ to match the size of $L\times L$. }
We show the meaning of other vectors in Figs. \ref{fig:adapter} (a1-a2). Note that \eqnref{eq:ad} only provides the scenario where RGB tokens are fused into depth in \figref{fig:adapter} (c). The pairwise fusion among other modalities follows similarly.

\begin{figure*}[t]
	\setlength{\abovecaptionskip}{2pt}
	\setlength{\belowcaptionskip}{-12pt}
 
	\centering 
	\includegraphics[width=0.9\textwidth]{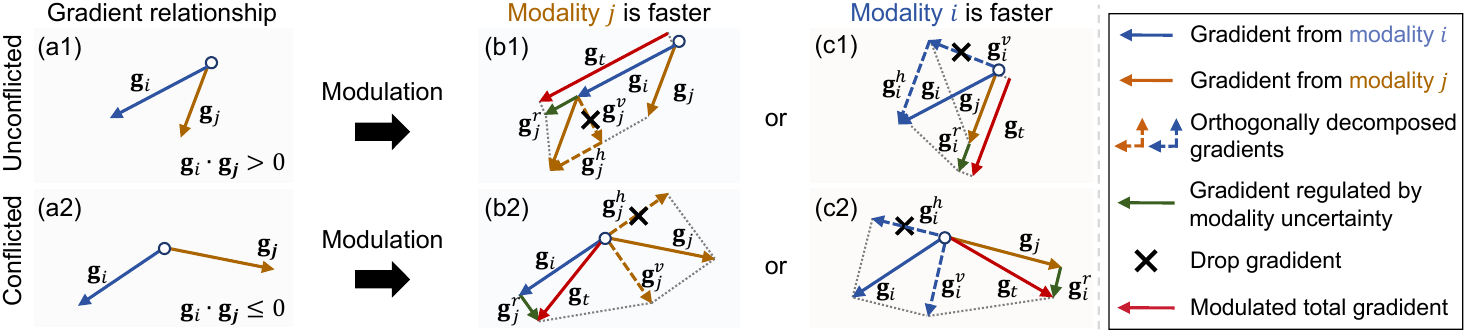}
    \caption{Illustration of gradient modulation via the proposed ReGrad in different scenarios: (Row 1) Non-conflicting (a1) and faster modality $j$ (b1) or $i$ (c1). (Row 2) Conflicting (a2) and faster modality $j$ (b2) or $i$ (c2).}
	\label{fig:grad} 
\end{figure*}

\vspace*{4pt}
\subsection{Rebalanced Modality Gradient Modulation}
\blue{Different modalities are suitable for distinct types of attacks and deployment conditions, \eg, the depth modality being less sensitive to light and more apt for detecting replay attacks, and the infrared modality excelling in identifying wearable mask attacks and performing well in low-light conditions.}
\blue{
To cope with various unknown attacks and deployment conditions, we aim to ensure that each modality can sufficiently converge, thereby fully leveraging their discriminative capability, rather than over-relying on a specific modality (\ie fast modality) that shows superior performance in the source domain.
}
Current modality balancing methods \cite{grad-cvpr22, pmr}, design for simple late fusion models, are limited in their applicability to FAS. 
Late fusion networks have been shown to underperform in FAS compared to hybrid fusion networks \cite{fm-vit,ma-vit} that allow the fusion both before the final classifier and during the feature extraction.
Additionally, the trainable parameters of a late fusion network are only affected by the gradient of its own modality, whereas those of a hybrid fusion network are influenced by the gradients of all modalities. Moreover, gradients from different modalities may even conflict with each other, making the modulation more difficult.
To this end, our ReGrad adaptively modulates the gradients of all modalities that simultaneously act on each trainable layer. 
Moreover, existing modality balancing methods fail to generalize well while balancing each modality due to the ignorance of domain shifts. 
Drawing inspiration from SSDG \cite{ssdg} and PMR \cite{pmr}, we introduce a single-side prototypical loss to supervise modality branches and monitor their convergence speed. This loss enables each modality to cluster according to domain information through internal impetus, which further mitigates the dominance of faster modalities.

\vspace*{4pt}
\noindent\textbf{Modulating Gradients for Each Modality.}
When the optimization direction is dominated by the gradients of the fast modality, the convergence degree of slow modalities is suppressed. 
The discriminative ability of slow modalities may even be weakened when there is a conflict in the gradient directions between modalities \cite{grad-multi-task}.
Inspired by the studies on solving domain conflicts \cite{grad-dg} and multi-task conflicts \cite{grad-multi-task} via gradient modulation, our ReGrad adaptively performs gradient modulation based on the conflict and convergence degree of each modality, making it applicable to complex hybrid fusion networks. 
\blue{
The criteria of modality convergence speed will be discussed in the following section.
}

As shown in \figref{fig:grad} (b1), if modality $i$ converges slower than $j$ and their gradients $\mathbf g_i$ and $\mathbf g_j$ are not conflicting, we suppress $\mathbf g_j$ and boost the learning of modality $i$.
This involves decomposing $\mathbf g_j$ into orthogonal $\mathbf g_j^v$ and non-orthogonal $\mathbf g_j^h$ components relative to $\mathbf g_i$.
We drop $\mathbf g_j^v$ to slow down modality $j$ and use $\mathbf g_j^h$ to accelerate modality $i$. Furthermore, to prevent unreliable modalities that converge rapidly from misleading other modalities, we employ the estimated uncertainty of modality $j$ to further constrain $g_j$.
Conversely, if $\mathbf g_i$ conflicts with $\mathbf g_j$ and modality $i$ remains a slow modal, it is unreasonable to use any part of $\mathbf g_j$ to boost the learning of $\mathbf g_i$.  To prevent $\mathbf g_j$ from seriously slowing down the convergence speed of modality $i$, as shown in \figref{fig:grad} (b2), we remove the non-orthogonal part of $\mathbf g_j^h$ that is opposite to $\mathbf g_i$. Meanwhile, we retain its orthogonal part $\mathbf g_j^v$ to allow modality $j$ to learn at a slower speed and reduce $\mathbf g_j$'s interference with $\mathbf g_i$. 
Similarly, the uncertainty-based suppression is adopted. Cases where the modality $i$ is faster are presented in Figs. \ref{fig:grad} (c1)-(c2) and \eqnref{eq:grad}.
\blue{
\begin{equation}
    \footnotesize
    \textrm{ReGrad}_2(\mathbf g_i,\! \mathbf g_j)\!=\!\left\{
    \begin{aligned}
        &\mathbf g_i + \frac{\mathbf g_i\cdot \mathbf g_j}{\|\mathbf g_i\|^2}\mathbf g_i\!\cdot\!\textrm U(u_j) & \!, & \,\textrm{in \figref{fig:grad} (b1)}, \\
        &\mathbf g_i + (\mathbf g_j - \frac{\mathbf g_i\cdot \mathbf g_j}{\|\mathbf g_i\|^2}\mathbf g_i)\!\cdot\!\textrm U(u_j) & \!, & \,\textrm{in \figref{fig:grad} (b2)}, \\
        &\frac{\mathbf g_i\!\cdot\!\mathbf g_j}{\|\mathbf g_j\|^2}\mathbf g_j\!\cdot\!\textrm  U(u_i) + \mathbf g_j & \!, & \,\textrm{in \figref{fig:grad} (c1)},\\
        &(\mathbf g_i - \frac{\mathbf g_i\cdot \mathbf g_j}{\|\mathbf g_j\|^2}\mathbf g_j)\!\cdot\!\textrm U(u_i) + \mathbf g_j & \!, & \,\textrm{in \figref{fig:grad} (c2)},
    \end{aligned}
    \right.
    \label{eq:grad}
\end{equation}
}
where $u_i$ and $u_j$ are respectively the batch-wise mean uncertainty of the class tokens outputted by the last ViT block of the corresponding modality, denoting the modality uncertainty.

\begin{table*}[t]
	\caption{Cross-dataset testing results under the fixed-modal scenarios (\textbf{Protocol 1}) among CASIA-CeFA (\textbf{C}), PADISI (\textbf{P}), CASIA-SURF (\textbf{S}), and WMCA (\textbf{W}). DG, MM, and FM are short for domain-generalized, multi-modal, and flexible-modal, respectively.}
        \vspace{-10pt}
	\setlength{\tabcolsep}{5pt}
        \renewcommand\arraystretch{0.78}
	\resizebox*{1.0 \linewidth}{!}{
		\begin{tabular}{lccccccccccc}
			\toprule[1.2pt]
			\multirow{2}{*}{\textbf{Method}} & \multirow{2}{*}{\textbf{Type}} & \multicolumn{2}{c}{\textbf{CPS $\rightarrow$ W}} &  \multicolumn{2}{c}{\textbf{CPW $\rightarrow$ S}} &  \multicolumn{2}{c}{\textbf{CSW $\rightarrow$ P}} &  \multicolumn{2}{c}{\textbf{PSW $\rightarrow$ C}} &  \multicolumn{2}{c}{\textbf{Average}} \\
			\cmidrule(lr){3-4}\cmidrule(lr){5-6}\cmidrule(lr){7-8}\cmidrule(lr){9-10}\cmidrule(lr){11-12}
			&  & HTER (\%) $\downarrow$ & AUC (\%) $\uparrow$ &  HTER (\%) $\downarrow$ & AUC (\%) $\uparrow$ &  HTER (\%) $\downarrow$ & AUC (\%) $\uparrow$ &  HTER (\%) $\downarrow$ & AUC (\%) $\uparrow$ & HTER (\%) $\downarrow$ & AUC (\%) $\uparrow$ \\
                \midrule[0.7pt]
			SSDG \cite{ssdg} & DG & 26.09  & 82.03  &  28.50  & 75.91  &  41.82  & 60.56  &  40.48  & 62.31  &  34.22 &        70.20  \\
			SSAN \cite{ssan} & DG & 17.73  & 91.69  &  27.94  & 79.04  &  34.49  & 68.85  &  36.43  & 69.29  &  29.15 &        77.22 \\
                IADG \cite{iadg} & DG & 27.02  & 86.50  &  23.04  & 83.11  &  32.06  & 73.83  &  39.24  & 63.68  &  30.34 &
                76.78 \\
                ViTAF \cite{adaptive-transformer} & DG & 20.58  & 85.82 & 29.16 & 77.80 & 30.75 & 73.03 & 39.75 & 63.44 & 30.06 & 75.02 \\
			MM-CDCN \cite{mm-cdcn} & MM & 38.92  & 65.39  &  42.93  & 59.79  & 41.38 & 61.51  &  48.14  & 53.71  &  42.84 & 60.10 \\
			CMFL \cite{cmfl} & MM & 18.22  & 88.82  & 31.20  & 75.66  &  26.68  & 80.85  & 36.93  & 66.82  & 28.26 & 78.04 \\
			ViT + AMA \cite{ama}& FM & 17.56  & 88.74  &  27.50  & 80.00  &  21.18  & 85.51  &  47.48  & 55.56  &  28.43 & 77.45 \\
			VP-FAS \cite{vp-fas}& FM & 16.26  & 91.22  &  24.42  & 81.07  &  21.76  & 85.46  &  39.35  & 66.55  &  25.45 & 81.08 \\
                \midrule[0.7pt]
                ViT \cite{vit} & Baseline & 20.88  & 84.77  &  44.05  & 57.94  &  33.58  & 71.80  &  42.15  & 56.45  &  35.16 & 67.74 \\
			\textbf{\modelName} & \textbf{Ours} & \textbf{12.79}  & \textbf{93.83}  &  \textbf{15.32}  & \textbf{92.86}  &  \textbf{18.95}  & \textbf{88.64}  &  \textbf{29.93}  & \textbf{76.52}  &  \textbf{19.25}  & \textbf{87.96} \\
			\bottomrule[1.2pt]
		\end{tabular}
	}
    \vspace*{-13pt}
    \label{tab:p1}
\end{table*}

\vspace*{4pt}
\noindent\textbf{Single-Side Prototypical Loss.}
Prototype learning has been successfully applied in FAS and modality rebalancing. PMR \cite{pmr} indicates that introducing prototypes to each modality in multi-modal learning to guide the learning process can mitigate modality imbalance.
However, vanilla prototype learning is suboptimal under DG scenarios as it neglects the differences in samples across various domains. Inspired by SSDG \cite{ssdg}, we propose the Single-Side Prototypical (SSP) loss with the concept of asymmetry representation and prototype learning. Specifically, we introduce prototypes for the live faces of all domains and the fake faces from each domain. Formulate the domain set as $\mathcal{D} = \{d_{\ell}, d_s^1, d_s^2, \dots ,d_s^{N_s}\}$, where $N_s$ is the number of source domains, $d_{\ell}$ denotes the domain consists of all live faces, and $d_s^i$ represents domain of spoof faces from the $i$-th dataset. For modality $m$, the prototype 
of domain $d\in\mathcal{D}$ is the center of all samples belonging to $d$ in the feature space: 
\begin{equation}
    \footnotesize
    p_{m}^d = \frac{1}{N_{d}}\sum_{c_m\in d}{c_m},
\end{equation}
where $c_m$ is a sample of $d$ in modality $m$, and $N_{d}$ denotes the total number of samples in $d$. 
Based on the prototypes of each domain, we aim to reduce the distance between samples and the prototype of their corresponding domain and increase the distance from other prototypes, thus preventing conflicts when samples from different domains are trained together.
Therefore, following the form of cross-entropy, the SSP loss $\mathcal{L}^{ssp}_m$ of modality $m$ is formulated as follows.
\begin{equation}
    \footnotesize
    \mathcal{L}_{m}^{ssp}(c_m, p^{d_{gt}}_m) = -\textrm{log}\frac{\textrm{exp}(-\textrm{ED}(c_m, p^{d_{gt}}_m))}{\sum_{d\in\mathcal{D}}{\textrm{exp}(-\textrm{ED}(c_m, p^{d}_m))}},
    \label{eq:ssp}
\end{equation}
where sample $c_m$ belongs to domain $d_{gt}$. The function $\textrm{ED}(\cdot)$ denotes the Euclidean distance of the input sample and its prototype. Our SSP loss encourages the features of each modality of the samples to be pulled to the corresponding domain's prototype. This optimization of each modality branch is driven by the intrinsic force within each modality. We mark the modality with lower $\mathcal{L}^{ssp}_{m}$ as a faster modality. Our final loss is expressed as follows.
\begin{equation}
    \footnotesize
    \label{eq:loss}
    \mathcal{L}^{final} = \mathcal{L}^{ce} + \sum_{m \in \{R, D, I\}}{\lambda \cdot \mathcal{L}^{ssp}_m},
\end{equation}
where $L^{ce}$ denotes the cross-entropy loss, and $\lambda$ is used to balance the trade-off between $\mathcal L^{ssp}_m$ and $\mathcal L^{ce}$. \blue{Here, we utilize $\mathcal L^{ce}$ to ensure the model's performance, while $\mathcal L^{ssp}_m$ is employed to balance all modalities and enhance generalizability.
Since $\mathcal L^{ssp}_m$ is not modality-specific, it does not undergo gradient modulation, which we perform to $\mathcal L^{ce}$ only.
}

\vspace{-2mm}
\section{Experimental Evaluation}
\subsection{Multi-Modal FAS Benchmark in DG Scenarios }
\label{sec:bench}
\noindent\textbf{Datasets, Protocols, and Performance Metrics.}
With the increasing popularity of large-scale foundation models \cite{sam}, the current trend is towards fully leveraging multiple and diverse source datasets/domains \cite{llm}. While training with multiple datasets has been widely studied for unimodal FAS \cite{ssdg, ssan, iadg}, the multi-modal FAS community still remains in the phase of single-dataset training \cite{fm-vit,ma-vit,cmfl}. Therefore, we propose the first large-scale multi-source benchmark to evaluate the DG performance of multi-modal FAS.
This benchmark incorporates four commonly used multi-modal datasets: CASIA-CeFA (\textbf{C}) \cite{cefa}, PADISI-Face (\textbf{P}) \cite{padisi}, CASIA-SURF (\textbf{S}) \cite{mmfa}, and WMCA (\textbf{W}) \cite{wmca}, each featuring a variety of attack types. For each dataset, we use RGB, depth (D), and infrared (I) modalities as inputs. Following previous works \cite{ssdg}, we use Half Total Error Rate (HTER) and Area Under the Receiver Operating Characteristic Curve (AUC) for evaluation.
We design three protocols to evaluate the cross-dataset performance under different unseen deployment conditions, \ie, fixed modalities, missing modalities, and limited source domains.
For \textbf{Protocol 1}, inspired by the leave-one-out (LOO) protocols designed in unimodal DG scenarios \cite{ssdg}, we propose four multi-modal LOO sub-protocols among \textbf{C}, \textbf{P}, \textbf{S}, and \textbf{W}. 
For example, the sub-protocol \textbf{CPS} $\rightarrow$ \textbf{W} represents we take \textbf{C}, \textbf{P}, and \textbf{S} as training sets, while \textbf{W} is testing set.
In \textbf{Protocol 2}, for each LOO sub-protocol of \textbf{Protocol 1}, we design three test-time missing-modal scenarios, \ie, D is missing, I is missing, and both D and I are missing. 
In \textbf{Protocol 3}, we limit the number of source domains by proposing two sub-protocols, namely \textbf{CW} $\rightarrow$ \textbf{PS} and \textbf{PS} $\rightarrow$ \textbf{CW}.
\noindent\textbf{Implementation Details.}
We resize all RGB, depth, and infrared images to $224\times224\times3$. Each modality's input is divided into $14\times14$ patch tokens and a class token is attached, collectively serving as the input to the ViT. 
The hidden size of each token is defaultly set to 768.
We employ the Adam optimizer with a learning rate of $5\times 10^{-5}$ and a weight decay of $1\times 10^{-3}$ to train 70 epochs. The batch size is set to 32. Our classifier is a single-layer fully-connected network that reduces the dimensionality of the class token outputted by the final ViT block from 768 to 2. 
The number $n$ of ViT blocks in MMDG is set to 12. 
ViT-b pretrained on ImageNet is adopted as the baseline. 

\begin{table*}[t]
    \caption{Cross-dataset testing results under the test-time missing-modal scenarios (\textbf{Protocol 2}) among CASIA-CeFA (C), PADISI (P), CASIA-SURF (S), and WMCA (W). We report the average HTER (\%) $\downarrow$ and AUC (\%) $\uparrow$ on four sub-protocols, \ie, \textbf{CPS $\rightarrow$ W}, \textbf{CPW $\rightarrow$ S}, \textbf{CSW $\rightarrow$ P}, and \textbf{PSW $\rightarrow$ C}. DG, MM, and FM are short for domain-generalized, multi-modal, and flexible-modal, respectively.}
    \vspace{-10pt}
    \centering
    \renewcommand\arraystretch{0.75}
    \setlength{\tabcolsep}{8.5pt}
    \small
    \resizebox*{1.0 \linewidth}{!}{
    \begin{tabular}{lccccccccc}
        \toprule[1.2pt]
        \multirow{2}{*}{\textbf{Method}} & \multirow{2}{*}{\textbf{Type}} & \multicolumn{2}{c}{\textbf{Missing D}} & \multicolumn{2}{c}{\textbf{Missing I}} & \multicolumn{2}{c}{\textbf{Missing D \& I}} & \multicolumn{2}{c}{\textbf{Average}} \\
        \cmidrule(lr){3-4}\cmidrule(lr){5-6}\cmidrule(lr){7-8}\cmidrule(lr){9-10}
         &  & HTER (\%) $\,\,\downarrow$ & AUC (\%) $\,\,\uparrow$ & HTER (\%) $\,\,\downarrow$ & AUC (\%) $\,\,\uparrow$ & HTER (\%) $\,\,\downarrow$ & AUC (\%) $\,\,\uparrow$ & HTER (\%) $\,\,\downarrow$ & AUC (\%) $\,\,\uparrow$ \\
        \midrule[0.7pt]
        SSDG \cite{ssdg} & DG & 38.92 & 65.45 & 37.64 & 66.57 & 39.18 & 65.22 & 38.58  & 65.75  \\
        SSAN \cite{ssan} & DG & 36.77 & 69.21 & 41.20 & 61.92 & 33.52 & 73.38 & 37.16  & 68.17  \\
        IADG \cite{iadg} & DG & 40.72 & 58.72 & 42.17 & 61.83 & 37.50 & 66.90 & 40.13 & 62.49  \\
        ViTAF \cite{adaptive-transformer} & DG & 34.99 & 73.22 & 35.88 & 69.40 & 35.89 & 69.61 & 35.59 & 70.64  \\
        MM-CDCN \cite{mm-cdcn} & MM & 44.90 & 55.35 & 43.60 & 58.38 & 44.54 & 55.08 & 44.35 & 56.27 \\
        CMFL \cite{cmfl} & MM & 31.37 & 74.62 & 30.55 & 75.42 & 31.89 & 74.29 & 31.27 & 74.78 \\
        ViT + AMA \cite{ama} & FM & 29.25 & 77.70 & 32.30 & 74.06 & 31.48 & 75.82 & 31.01 & 75.86 \\
        VP-FAS \cite{vp-fas} & FM & 29.13 & 78.27 & 29.63 & 77.51 & 30.47 & 76.31 & 29.74 & 77.36 \\
        \midrule[0.7pt]
        ViT & Baseline & 40.04 & 64.69 & 36.77 & 68.19 & 36.20 & 69.02 & 37.67 & 67.30\\
        \textbf{MMDG} & \textbf{Ours} & \textbf{24.89} & \textbf{82.39} & \textbf{23.39} & \textbf{83.82} & \textbf{25.26} & \textbf{81.86} & \textbf{24.51} & \textbf{82.69} \\
        \bottomrule[1.2pt]
    \end{tabular}
    }
    \vspace*{-10pt}
    \label{tab:p2}
\end{table*}

\subsection{Cross-dataset Testing}
We compare our method on \textbf{Protocols 1-3} with three categories of FAS methods: (1) Multi-modal methods, including CMFL \cite{cmfl} and MM-CDCN \cite{mm-cdcn}. (2) Flexible-modal FAS, such as ViT + AMA \cite{ama} and VP-FAS \cite{vp-fas}. (3) Domain generalized FAS \ie, SSDG \cite{ssdg}, SSAN \cite{ssan}, IADG \cite{iadg}, and ViTAF \cite{adaptive-transformer}. 
For methods not designed for unimodal FAS, we re-implement them either by employing a concatenation of multi-modal inputs or by replacing the backbone with our ViT+\adName~or vanilla ViT. All these methods are trained on the proposed protocols. We automatically use their hyper-parameters as described in the corresponding reference papers or optimally assign better ones. 

\vspace*{4pt}
\noindent\textbf{Protocol 1: Fixed Modalities.}
We evaluated the cross-dataset performance of unimodal DG and multi-modal methods in the fixed modalities scenarios. Since unimodal DG methods cannot be directly applied to this protocol, we concatenate the images of the three modalities along the channel dimension. We then reduce the channel number to 3 using a trainable $1\times1$ convolution layer before feeding them to these methods.
As shown in the \tabref{tab:p1}, our \modelName~achieves the best results. Compared to the second-ranked method, \ie VP-FAS \cite{vp-fas}, there is an improvement of over 6\% in both HTER and AUC. A higher performance boost occurs when compared to domain-generalized methods.
This indicates \modelName 's effectiveness in addressing modality unreliability and imbalance.

\vspace*{4pt}
\noindent\textbf{Protocol 2: Missing Modalities.}
As shown in \tabref{tab:p2}, our MMDG outperforms other algorithms in all three scenarios of modality missing during testing. It's worth noting that ViT+AMA and VP-FAS are specifically designed for scenarios of modality missing and have achieved better results compared to other methods. However, under the DG premise, they did not consider the modality imbalance and unreliability issues brought by domain shift, resulting in a difference of over 5\% in average HTER and AUC compared to our method. 
Another interesting observation is that SSAN, IADG, ViT, and ViT + AMA perform better in the \textbf{Missing D \& I} scenario than in \textbf{Missing D} and \textbf{Missing I}. These methods lose more modalities but achieve performance improvement, indicating that unreliable information within the modalities leads to erroneous detection. This reflects the effectiveness of our approach in suppressing the interaction of unreliable information between modalities.

\begin{table}[t]
\setlength{\tabcolsep}{5pt}
\caption{Cross-dataset testing results under the limited source domain scenarios (\textbf{Protocol 3}) among CeFA-CeFA (\textbf{C}), PADISI-USC (\textbf{P}), CASIA-SURF (\textbf{S}), and WMCA (\textbf{W}).}
\vspace{-10pt}
\setlength{\tabcolsep}{7.9pt}
\renewcommand\arraystretch{0.8}
\resizebox*{1.0 \linewidth}{!}{
    \begin{tabular}{lccccc}
    \toprule[1.2pt]
    \multirow{2}{*}{\textbf{Method}} & \multirow{2}{*}{\textbf{Type}} & \multicolumn{2}{c}{\textbf{CW} $\rightarrow$ \textbf{PS}} & \multicolumn{2}{c}{\textbf{PS}$\rightarrow$ \textbf{CW}} \\
    \cmidrule(lr){3-4}\cmidrule(lr){5-6}
     &  & HTER (\%) & AUC (\%) & HTER (\%) & AUC (\%) \\
    \midrule[0.7pt]
    SSDG \cite{ssdg} & DG & 25.34 & 80.17 & 46.98 & 54.29   \\
    SSAN \cite{ssan} & DG & 26.55 & 80.06 & 39.10 & 67.19   \\
    IADG \cite{iadg} & DG & 22.82 & 83.85 & 39.70 &  63.46  \\
    ViTAF \cite{adaptive-transformer} & DG & 29.64 & 77.36 & 39.93 & 61.31 \\
    MM-CDCN \cite{mm-cdcn} & MM & 29.28 & 76.88 & 47.00 &  51.94  \\
    CMFL \cite{cmfl} & MM & 31.86 & 72.75 & 39.43 & 63.17 \\
    ViT + AMA \cite{ama} & FM & 29.25 & 76.89 & 38.06 & 67.64 \\
    VP-FAS \cite{vp-fas} & FM & 25.90 & 81.79 & 44.37 & 60.83 \\
    \midrule[0.7pt]
    ViT \cite{vit} & Baseline & 42.66 & 57.80 & 42.75 & 60.41 \\
    \textbf{\modelName} & \textbf{Ours} & \textbf{20.12} & \textbf{88.24} & \textbf{36.60} & \textbf{70.35} \\
    \bottomrule[1.2pt]
    \end{tabular}
}
\label{tab:p4}
\vspace{-1pt}
\end{table}

\vspace*{4pt}
\noindent\textbf{Protocol 3: Limited Source Domains.}
Compared to protocol 1, the limitation of source domains results in a more significant domain shift. We observe that our proposed MMDG achieves optimal results in both sub-protocols (see \tabref{tab:p4}), even in the presence of substantial domain shift in \textbf{PS} $\rightarrow$ \textbf{CW}. This further underscores the effectiveness of our \adName~and \gradName~in enhancing domain generalizability in unseen multi-modal deployment environments.

\subsection{Ablation Study and Discussion}
In this section, we perform ablation analysis on \adName~and \gradName~to verify their individual contribution.

\noindent\textbf{Effectiveness of \adName.}
\begin{figure}[t]
	\setlength{\abovecaptionskip}{2pt}
	\setlength{\belowcaptionskip}{-0.5mm}
	\centering 
	\includegraphics[width=0.48\textwidth]{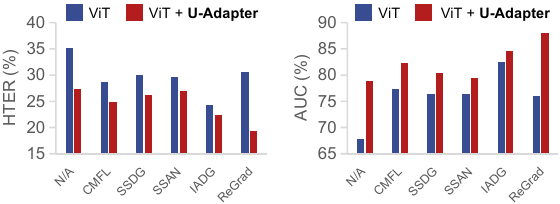}
        \vspace{-5mm}
	\caption{Ablation results on our \adName. We report average HTER $\downarrow$ and AUC $\uparrow$ on four sub-protocols in \textbf{Protocol 1}.} 
	\label{fig:abla_adapter} 
 \vspace{-1mm}
\end{figure}
To validate the effectiveness of the proposed \adName, we compare the network performance before and after removing the adapter. We also replace the backbone of existing DG methods (\ie, SSDG, SSAN, and IADG) and multi-modal methods (CMFL) to observe the performance differences when using ViT+\adName~as the backbone versus using ViT alone. 
As shown in \figref{fig:abla_adapter}, compared to using ViT as the backbone, these methods experience a performance boost when additional \adName s are employed. The performance improvement is especially pronounced when no additional method is used, and when our proposed \gradName~is integrated. This demonstrates the effectiveness of \adName~and its universality for existing DG methods and multi-modal strategies.

\begin{table}[t]
\caption{Ablation results on the proposed \gradName. We report the average HTER and AUC on four sub-protocols in \textbf{Protocol 1}.}
\vspace{-10pt}
\renewcommand\arraystretch{0.79}
\resizebox*{1.0 \linewidth}{!}{
\begin{tabular}{cccc}
    \toprule[1.2pt]
    \textbf{Backbone} & \textbf{Method} & \textbf{HTER} (\%) & \textbf{AUC} (\%) \\
    \midrule[1.0pt]
    ViT \cite{vit} & -  & 31.14 & 74.81 \\
    ViT \cite{vit} & \textbf{\gradName} & \textbf{30.59} & \textbf{75.94} \\
    \midrule[0.5pt]
    ViT + \adName & SSDG \cite{ssdg} & 26.13 & 80.31 \\
    ViT + \adName & SSAN \cite{ssan} & \multicolumn{1}{c}{26.94} & \multicolumn{1}{c}{79.52} \\
    ViT + \adName & IADG \cite{iadg} & \multicolumn{1}{c}{24.23} & \multicolumn{1}{c}{82.54} \\
    ViT + \adName & PMR \cite{pmr} & \multicolumn{1}{c}{24.54} & \multicolumn{1}{c}{83.14} \\
    ViT + \adName & OGM-GE \cite{grad-cvpr22} & \multicolumn{1}{c}{29.32} & \multicolumn{1}{c}{76.54} \\
    ViT + \adName & - & 27.38 & 78.88 \\
    ViT + \adName & \textbf{\gradName} (Conflicted-only) & 21.91 & 85.65 \\
    ViT + \adName & \textbf{\gradName} (Unconflicted-only) & 21.23 & 85.51 \\
    ViT + \adName & \textbf{\gradName}  & \textbf{19.48} & \textbf{87.48} \\
    \bottomrule[1.2pt]
    \end{tabular}
}
\label{tab:abla_grad}
\vspace{-13pt}
\end{table}

\begin{figure}[t]
	\setlength{\abovecaptionskip}{2pt}
	\setlength{\belowcaptionskip}{-2pt}
	\centering 
	\includegraphics[width=0.46\textwidth]{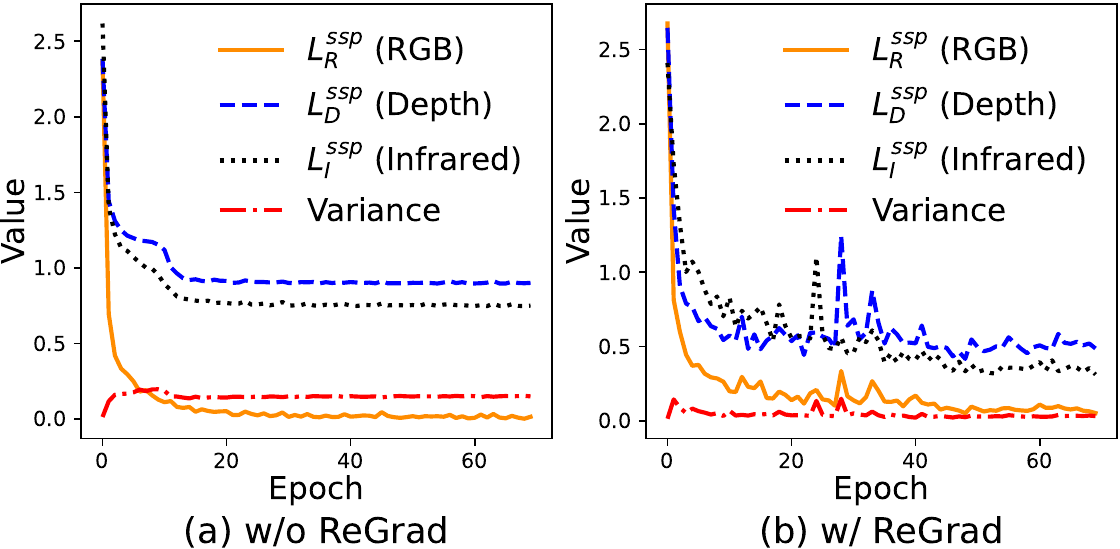}
	\caption{Modality balance degree during training. (a) MMDG w/o ReGrad. (b) MMDG w/ ReGrad. \blue{The red dashed line in each sub-figure represents the variance of $\mathcal L^{ssp}_R$, $\mathcal L^{ssp}_D$, and $\mathcal L^{ssp}_I$.}} 
	\label{fig:curve} 
 \vspace{-0.1mm}
\end{figure}

\vspace*{4pt}
\noindent\textbf{Effectiveness of \gradName.}
Here, we try to answer two questions: (1) \textit{Whether the idea to address domain shift through modality rebalancing, is more suitable compared to existing DG methods?}
(2) \textit{Is our ReGrad more capable of enhancing the performance of complex cross-modal fusion frameworks like ViT+\adName~compared to existing modality balancing methods?}
From \tabref{tab:abla_grad}, we can observe that when ViT+\adName~is used as the backbone, our \gradName~surpasses existing DG methods and other modality rebalancing methods. Even for vanilla ViT, our \gradName~can enhance its performance. This indicates the effectiveness of addressing domain shifts in unseen scenarios by balancing various modalities. The results in \tabref{tab:abla_grad} also indicate that when compared to the SoTA modality balancing methods (\ie, OGM-GE and PMR), our \gradName~is more suitable for ViT + \adName, which involves hybrid feature fusion and complex cross-modal interactions. 

Meanwhile, as illustrated in \figref{fig:curve}, the lower variance (right) of $\mathcal L^{ssp}_{R}, \mathcal L^{ssp}_{D}$, and $\mathcal L^{ssp}_{I}$ demonstrates that \gradName~obviously improves the balance degree of MMDG. We also notice that ReGrad causes slight fluctuations in the loss function values. 
This is due to the gradient modulation allowing the model to escape from local optima, seeking a direction with a higher degree of modality balance. Despite the fluctuations, MMDG can still converge after about 50 epochs.
Additionally, we compare \gradName~with its two variants, which perform modulation only when modality gradients are conflicted and unconflicted, respectively. Although these two variants improve performance compared to not modulating gradients at all, their performance is inferior to modulating gradients in all situations. This validates the necessity of adaptive modulation under various scenarios.

\begin{table}[t]
\caption{Impact of estimated uncertainty in \adName s and \gradName. We report the average HTER (\%) $\downarrow$ and AUC (\%) $\uparrow$ on four sub-protocols in \textbf{Protocol 1}.}
\vspace{-10pt}
\renewcommand\arraystretch{0.4}
\resizebox*{1.0 \linewidth}{!}{
    \begin{tabular}{cccc}
    \toprule[1.0pt]
    \multicolumn{2}{c}{\textbf{Using Estimated Uncertainty}} & \multicolumn{2}{c}{\textbf{Metric Value}} \\
    \cmidrule(lr){1-2}\cmidrule(lr){3-4}
    \adName: \eqnref{eq:ad} & \gradName: \eqnref{eq:grad} & HTER (\%) $\downarrow$ & AUC (\%) $\uparrow$ \\
    \midrule[1.0pt]
     \tiny\XSolid & \tiny\XSolid & 28.40 & 78.65 \\
     \tiny\XSolid & \small\Checkmark & 26.13 & 81.60 \\
     \small\Checkmark & \tiny\XSolid & 22.75 & 84.84 \\
     \small\Checkmark & \small\Checkmark & \textbf{19.48} & \textbf{87.48} \\
    \bottomrule[1.0pt]
    \end{tabular}
}
\label{tab:abla_uc}
\vspace{-8pt}
\end{table}

\begin{figure}[t]
	\setlength{\abovecaptionskip}{5pt}
	\setlength{\belowcaptionskip}{-2pt}
	\centering 
	\includegraphics[width=0.47\textwidth]{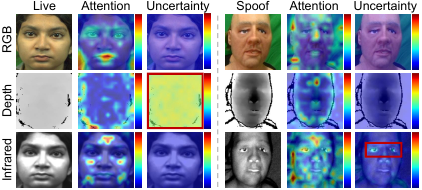}
	\caption{Visual attention and uncertainty maps of our MMDG. Uncertainly predicted regions are highlighted with red boxes. \blue{The redder color indicates the higher uncertainty or more attention.}} 
	\label{fig:attention} 
\end{figure}
\vspace*{4pt}
\noindent\textbf{Impact of the Uncertainty in \adName~and \gradName.}
To testify whether the estimated uncertainty can enhance detection performance, we made a comparison by removing the parts that use uncertainty for constraints in \adName~and \gradName. From \tabref{tab:abla_uc}, it can be observed that adding uncertainty to constrain the propagation of unreliable information in \adName~and the gradient of fast modalities in \gradName~both lead to performance improvement. When uncertainty is added to both modules, HTER and AUC can be further improved by over 3\%.
\blue{
As shown in \figref{fig:attention}, MMDG can identify unreliable regions in depth (left) and infrared (right) modalities.
Although the U-Adapters of these modalities focus on regions with uncertain predictions, due to our idea of suppressing uncertain regions, these regions fail to mislead mutual modalities, enabling the capture of more reliable spoof traces against domain shifts.
}
These observations validate the effectiveness of using uncertainty to constrain unreliable modalities.

\vspace*{4pt}
\noindent\textbf{Discussion.}
The results presented in Tables \ref{tab:p1}-\ref{tab:abla_uc} and Figs. \ref{fig:abla_adapter}-\ref{fig:curve} validate the MMDG's generalizability by addressing the problems of modality unreliability and imbalance. 
This work marks a stride towards practical deployments and can benefit the FAS community. However, our benchmark is new and MMDG's performance can still be improved. Continued efforts on this benchmark are necessary for further progress. Meanwhile, our method also has space for improvement, \eg, further enhancing the degree of modality balance, preventing the desecration of training stability, and adopting more advanced uncertainty estimation techniques to prevent the risk of the model's over-confidence.

\section{Conclusion}
In this work, we propose the first multi-modal DG framework, \ie \modelName, for FAS to enhance generalizability by addressing modality unreliability and imbalance.
To mitigate the interference of unreliable modalities, we propose \adName~to suppress uncertainly predicted regions during cross-modal feature fusion. Besides, we develop \gradName~to adaptively modulate the gradients of each modality branch to achieve modality balance, thus preventing an over-reliance on a dominant modality.
\blue{We also establish the first large-scale} benchmark for multi-modal FAS under DG scenarios. Extensive experiments on this benchmark demonstrate \modelName 's effectiveness.

\vspace*{4pt}
\textbf{Acknowledgement.} This work was supported by the National Key Research and Development Program of China under Grant no.2022YFB3207700. The support funding was also from the National Natural Science Foundation of China under Grant 62272022, 62306061, and U22A2009.

\newpage
{
    \small
    \bibliographystyle{ieeenat_fullname}
    \bibliography{main}
}

\end{document}